\documentclass{article}

\usepackage{microtype}
\usepackage{verbatim}
\usepackage{graphicx}
\usepackage{subfig}
\usepackage{multirow}
\usepackage{booktabs}
\usepackage{hyperref}
\usepackage{caption}

\usepackage[accepted]{sysml2019}
\sysmltitlerunning{Hardware Aware Neural Network Architectures}
\begin{document}

\twocolumn[
\sysmltitle{Hardware Aware Neural Network Architectures (using FBNet)}.
\sysmlsetsymbol{equal}{*}

\begin{sysmlauthorlist}
\sysmlauthor{Sai Vineeth Kalluru Srinivas}{}
\sysmlauthor{Harideep Nair}{}
\sysmlauthor{Vinay Vidyasagar}{}
\end{sysmlauthorlist}

\begin{center}
\textbf{Carnegie Mellon University - Silicon Valley}
\end{center}

\sysmlkeywords{Neural Architecture Search, MobileNetV2, CondenseNet, FBNets, Hardware Architecture}

\vskip 0.3in

\begin{abstract}
We implement a differentiable Neural Architecture Search (NAS) method inspired by FBNet for discovering neural networks that are heavily optimized for a particular target device. The FBNet NAS method discovers a neural network from a given search space by optimizing over a loss function which accounts for accuracy and target device latency. We extend this loss function by adding an energy term. This will potentially enhance the ``hardware awareness" and help us find a neural network architecture that is optimal in terms of accuracy, latency and energy consumption, given a target device (Raspberry Pi in our case). We name our trained child architecture obtained at the end of search process as Hardware Aware Neural Network Architecture (HANNA). We prove the efficacy of our approach by benchmarking HANNA against two other state-of-the-art neural networks designed for mobile/embedded applications, namely MobileNetv2 and CondenseNet for CIFAR-10 dataset. Our results show that HANNA provides a speedup of about 2.5x and 1.7x, and reduces energy consumption by 3.8x and 2x compared to MobileNetv2 and CondenseNet respectively. HANNA is able to provide such significant speedup and energy efficiency benefits over the state-of-the-art baselines at the cost of a tolerable 4-5\% drop in accuracy.

\end{abstract}
] 

\section{INTRODUCTION}
\label{intro}
Deep Learning research has been gaining ever-increasing impetus, especially over the past few years. Every year, around five to ten new state-of-the-art architectures are proposed which surpass human-level performance. However, all these architectures mainly focus on improving accuracy but at the cost of increased model complexity. Exploding model size puts pressure on hardware resources in terms of performance, area and energy consumption. Some models take up to four to eight seconds during inference stage. The latency during inferencing would be crucial for real-time applications like self-driving cars, air traffic control and others. Further, these models consume a lot of energy with increasing layers of depth. The current state-of-the-art models are not designed for applications with low-energy hardware constraints. Hence, recently there has been growing interest in training neural architectures that could discover child neural networks (childnets) which are better optimized for the target hardware device. We implement one such method called Differentiable Neural Architecture Search (NAS) and use an approach inspired by FBNet NAS. 

\subsection{Need for Hardware Aware Neural Networks}

Neural networks research has been exploding at an exponential rate. The current state-of-the-art models like Inception-v4, VGG-16, ResNet-50 have outperformed other neural networks for computer vision related tasks with accuracy of prediction at par or sometimes better than human-level performance. The depth of the current networks range anywhere between 15 to 150 layers. This depth allows them to obtain very high accuracy for the related task. However, every layer requires heavy computation and millions of matrix multiplications. Such heavy computations take long time for execution on devices that do not possess the high compute resources required by the model. Typically mobile and embedded devices come with hardware constraints like low GFLOPs/sec or memory bandwidth which would make the current models run very slow during the inference stage, making them impractical for real-time applications. Another impeding constraint is energy consumption since such devices typically operate on batteries or low power.

This brings the need to optimize well performing neural networks for a required application on a target device. Specifically the optimized neural network should have low-latency and low-energy consumption on the particular target device. Traditionally, this problem is solved by using methods that involve hand engineering like pruning, compression, quantization, fixed-width representation among others. Each of these methods is not very intuitive and requires highly skilled ML engineers with subject matter expertise. The models are iteratively improved which is time consuming. Most of these optimizations do not exploit the full potential of the hardware. 
\begin{figure}
    \centering
    \includegraphics[width=0.35\textwidth]{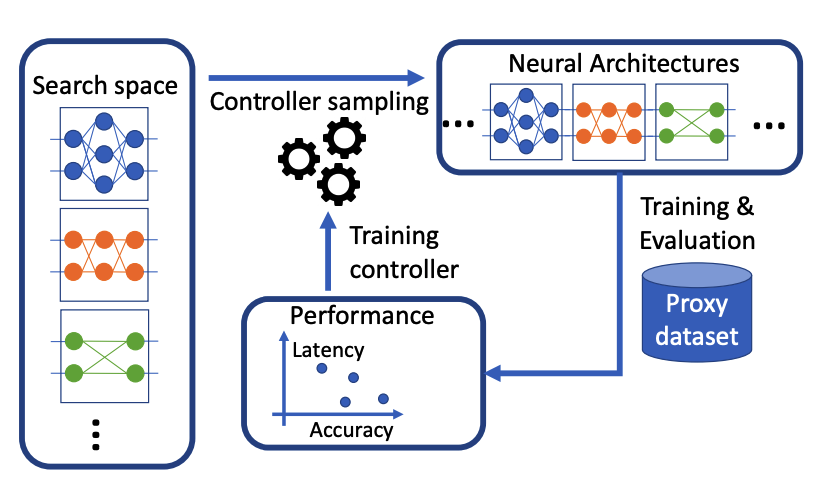}
    \caption{Block diagram showing the process of optimizing a neural network for a target device. The gears symbol shows a \textit{controller} architecture which automatically optimizes over a given search space \cite{wu2018fbnet}}
    \label{fig:nas_block_diagram}
\end{figure}

\subsection{Neural Architecture Search}

To tackle the above mentioned problem, in recent times many research groups have proposed a method that discovers a neural architecture over a search space by optimizing a loss function which embeds some form of accuracy along with latency introduced (by a particular model) on the target device. Fig. \ref{fig:nas_block_diagram} shows the working of such a method. Typically a \textit{controller} is employed which samples a neural network from a given search space, trains it with a proxy dataset followed by calculating its latency and accuracy values. Based on its performance the \textit{controller} picks a different architecture from the search space that would try to minimize latency while maintaining similar accuracy to the original high accuracy one. The controller architecture is usually a super net architecture that discovers a child architecture that is optimized.

\section{Related Work}
\label{related_work}
Neural Architecture Search works as the controller architecture with the primary task of optimizing neural network for a target device. A lot of work has been published in this area, and we would discuss few prominent ones. 

\subsection{MobileNASNet}
MobileNASNet \cite{tan2018mnas} is a NAS method specially focused on optimizing models for mobile devices. Briefly, it employs a gradient-based Reinforcement Learning method to find solution for a multi-objective problem. The search framework consists of three components: a recurrent neural network (RNN) based controller, a trainer to obtain the model accuracy, and a mobile phone based inference engine for measuring latency. This model makes use of simplified cell-level search to find out the right blocks for a prefixed child net \textit{macro-architecture}. Even though cell-level searching works pretty well, implementing the cell-level child architecture on a hardware is quite complex.

\subsection{DARTS - Differentiable Architecture Search}

\begin{figure}
    \centering
    \includegraphics[width=0.35\textwidth]{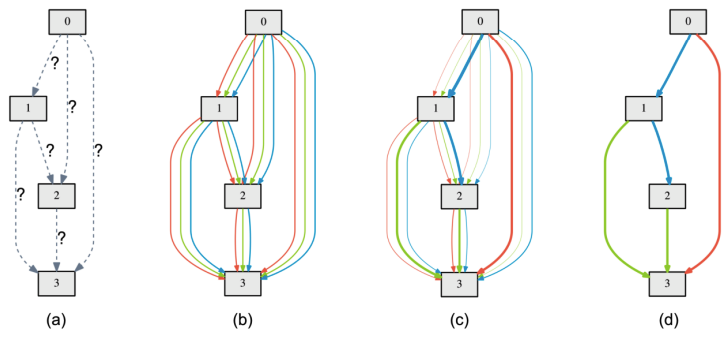}
    \caption{An overview of DARTS: (a) Operations on the edges are initially unknown. (b) Continuous
relaxation of the search space by placing a mixture of candidate operations on each edge. (c) Joint
optimization of the  probabilities and the network weights 
problem. (d) Deriving the final architecture from the learned  probabilities\cite{liu2018darts}.}
    \label{fig:darts}
\end{figure}

The DARTS \cite{liu2018darts} method employs differentiable search methods to discover child networks over a \textit{continuous} search space. This method makes the search space continuous by introducing a simple relaxation scheme as shown in Fig. \ref{fig:darts}. Briefly explained, the blocks in the figure could be Conv, Pool, ReLU blocks etc. The edges (dataflow) between these blocks are initially unknown. Later, all possibilities of connections are explored and optimized. The final architecture is inferred from learned probabilities.

\subsection{MONAS - Multi-objective NAS}

MONAS \cite{hsu2018monas} is another effective NAS method developed for a multi-objective requirement. MONAS also uses Reinforcement Learning (RL) by using a reward function. The accuracy and energy consumption of the target network are the rewards to the robot network. An RNN is used as super net that supplies hyperparameters to a CNN based child net architecture. The robot network updates itself based on this reward with RL.   

\section{HANNA (based on FBNet)}\label{fbnet}
\begin{figure}[!hbt]
    \centering
    \includegraphics[width=0.4\textwidth]{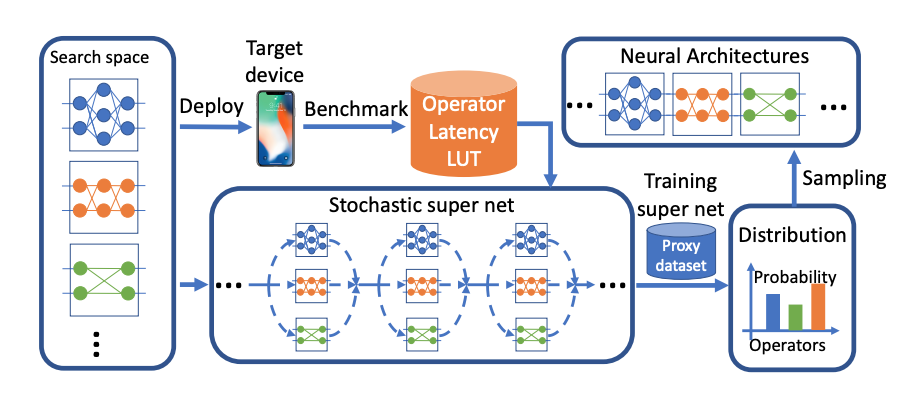}
    \caption{FBNet Search process\cite{wu2018fbnet}}
    \label{fbNetSearch}
\end{figure}

FBNets\cite{wu2018fbnet} are optimal ConvNet architectures that are obtained using DNAS (Differentiable Neural Architecture Search). The term FBNet refers to a set of child architectures, chosen by a DNAS based approach, that are efficient and accurate to run on resource constrained platforms like mobile phones. Fig.\ref{fbNetSearch} shown below portrays how FBNets are obtained. The super net trains the child networks in a stochastic manner and once this is done the networks are sampled based on a distribution. The optimal sampled network is the network that is used on the target device.
 ConvNets used in Computer Vision are plagued with problems related to  design space issues, lack of standardized efficiency metrics, and non transferable optimality. The design space is very large in the case of ConvNets and thus it makes it harder to tweak these models. There is a lack of a standardized efficiency metric on hardware platforms. The lack of consistency between the software development of ConvNets and the hardware constraints makes it a hard problem to solve. Further, the network can't be automatically tuned on a target platform and given the design time the process seems unfeasible.
 
\begin{figure}[!hbt]
     \centering
     \subfloat[MobileNetv2 based candidate block]{\includegraphics[width=0.2\textwidth]{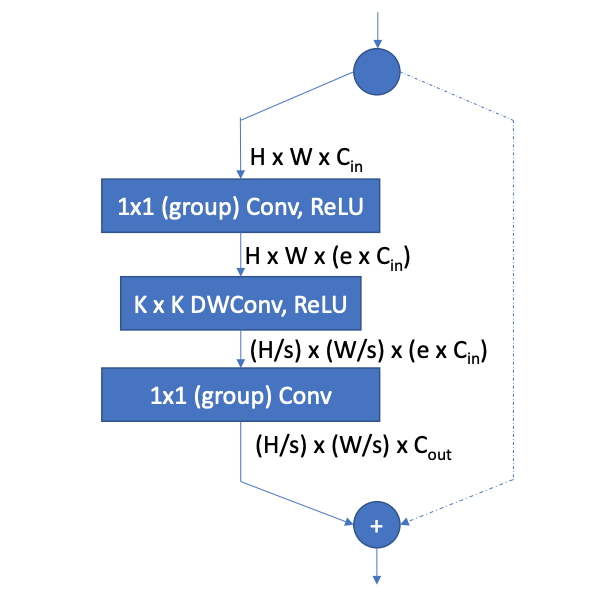}}
     \qquad
     \subfloat[Configurations for 9 candidate blocks]{\includegraphics[width=0.24\textwidth]{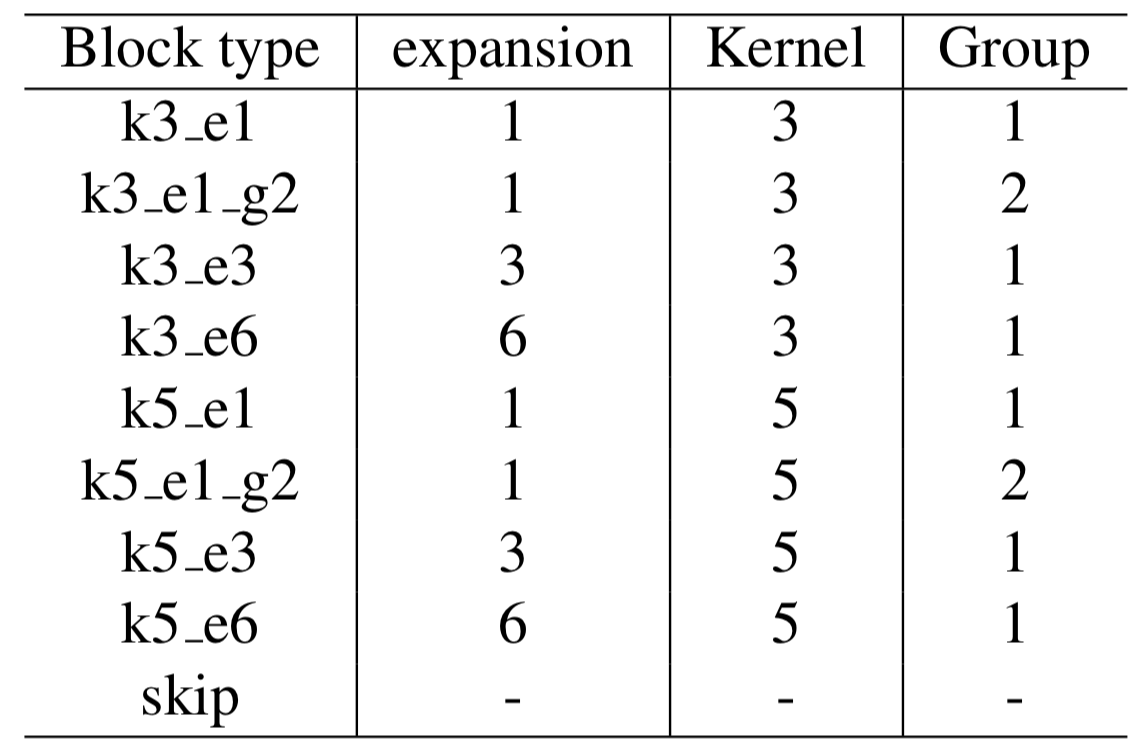}}
     \caption{Candidate blocks \cite{wu2018fbnet}}
     \label{blockFBnet}
 \end{figure}

DNAS allows a layer wise block search rather than a cell level hierarchical search making it easier to search for Conv models over a large search space. The search space is represented as stochastic super net that is trained by optimizing a particular loss function.
 \begin{equation}
     min_{ a \in A} min_{w_{a}} L(a,w_{a})
     \label{LossMinFbnet}
 \end{equation}
 Equation \ref{LossMinFbnet} expresses the neural architecture search problem. The NAS will try to find the weights that minimize the loss given a particular architecture "a". The search is performed at a block level where each layer can search through a set of nine blocks, that provide the best accuracy and minimize the loss. Thus, there are ${9^{22}}$ possibilities, searching through which is not a trivial task. The candidate blocks are derived from MobileNetv2 \cite{sandler2018mobilenetv2}. Fig.\ref{blockFBnet}(a) shows the architecture of such a block, consisting of depthwise separable convolution and linear bottleneck. Fig. \ref{blockFBnet}(b) shows the 9 different configurations of the block with various expansion ratios, kernel and group sizes.
 \begin{figure}[!hbt]
    \centering
    \includegraphics[width=0.35\textwidth]{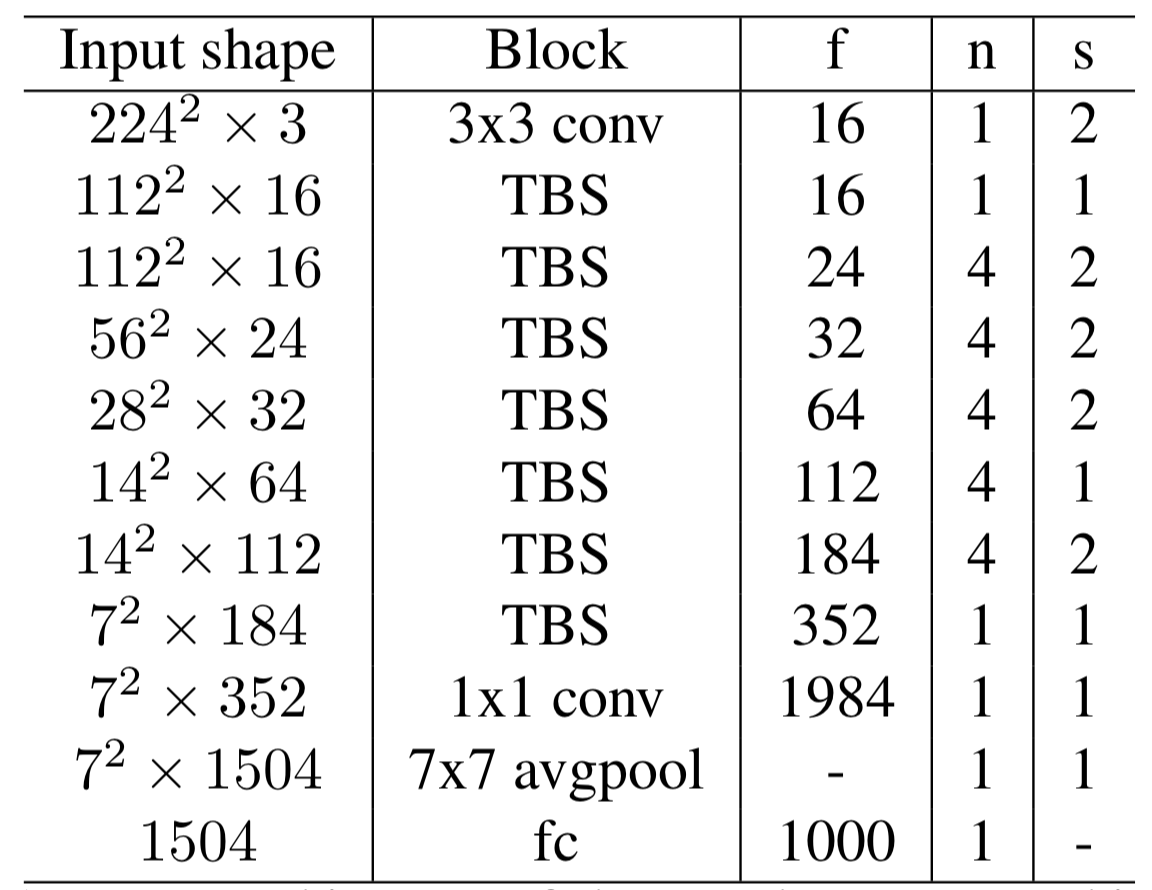}
    \caption{Fixed macro-architecture of FBNet\cite{wu2018fbnet}}
    \label{macroarch}
\end{figure}
 
 To find optimal blocks per layer, as mentioned before, the loss function needs to take into account accuracy, latency and energy. Equation \ref{newOptimization} below shows the loss to be optimized. Please note that the FBNet authors only proposed a two-term loss function that does not take into account energy consumption.
$\alpha$ and $\beta$ allow us to tune the penalty incurred from latency of the childnet, while $\gamma$ and $\delta$ allow us to tune the penalty incurred due to energy of the childnet.
 \begin{equation}
 L(a,w_a) = CE(a,w_a) + \alpha*LAT(a)^\beta + \gamma*ENER(a)^\delta
\label{newOptimization}
\end{equation}
 
Architectures can be sampled with a probability of $P(\theta)$. $\theta$ refers to an architecture sampling parameter modeled by a Softmax function, which represents the probability of each of the 9 candidate blocks to be in each of the 22 TBS layers. Given the high-dimensional search space, the optimization problem can be written as in (\ref{newOptimi}).
 \begin{equation}
     min_\theta min_{wa} E_{a~P(\theta)}{L(a,w_a)}
     \label{newOptimi}
 \end{equation}
(\ref{newOptimi}) offers an optimization by not solving for the optimal architecture "a" but rather optimizing $P(\theta)$ of the super net to minimize the expected loss. Further, by using a Gumbel-Softmax we can introduce continuity in our sampling parameter. The latency term can be written as shown in (\ref{updatedLAT}).
\begin{equation}
    LAT(a)=\sum_l\sum_i m_{li}.LAT(b_{li})
    \label{updatedLAT}
\end{equation}
(\ref{updatedLAT}) shows an important result as the latency of an architecture can be expressed as the product of a mask$(m_{li})$, that is expressed as Gumbel-Softmax function, which contains $\theta$, and the latency across a block in a layer. A similar approach is used for energy. This allows latency and energy to be differentiable. Thus the super net can optimize both accuracy through weights, and latency and energy through $\theta$.

\section{Achieved Milestones}
\subsection{Profiling on Raspberry Pi}

Profiling the Raspberry Pi for latency and energy metrics is an essential step. This process helps us get real world behavior of the model on the target device. Using profiled latency and energy we would be able to fine tune a model such that it complies with the target devices limits. 

A latency look up table was used as crucial component in FBNet NAS architecture. The loss functions penalises the weights of the model according to the sum of latencies produced by each candidate block for child architecture in the current iteration. 
The look up table was generated by finding the execution time of a particular candidate block within our search space. The execution time was written to a file that served as our look up table.

Energy was the second profiling metric that was utilized in our optimization. The factors that affect energy are Voltage, Current, and time. Voltage was fixed at the supply voltage of RPi, at about 5.1V. Current consumed on the other hand varied based on the computation - heavier the computation the more would be the current draw. The current draw was measured using a PowerJive stick connected to the RPi.


The idle current draw of RPi was measured to be around 0.24A. Under inference workload the RPi drew a current between a range of 0.24 to 0.74A.  It was observed that RPi's current draw followed a peak and dip pattern. The current increased from 0.24A to normally within 0.35A to a max value of 0.73A(based on the block) and then fell down to 0.24A after inference.The current values that were observed during inference time consisted both of static and dynamic current. The current value that was captured for our look up table was the difference between the peak current/dynamic current drawn and the static current. The noted current value in lookup table formed the basis of energy calculation.

Initially,for energy consumed we used a fixed value of time taken per candidate block but this proved to give us inaccurate models. The modified energy calculation took into account the current draw per block and the actual time per block based on the generated latency look up table. 

We also tried to consider just power rather than energy but this produced ineffective models in terms of minimizing latency and power consumed. Therefore, we stuck to using energy in our optimization equation.

\subsection{Baseline Calculations}
Earlier section talks about how profiling was performed on our target device Raspberry Pi 3B. It comes with a 1.4GHZ quad-core ARM processor offered as SoC by Broadcom. RPi is popular among hobbyists for quick prototyping.

In the literature of NAS architectures, FBNet stands special in its objective of performing multi-objective optimization for ARM devices in particular. The results of FBNet paper claim to achieve better accuracy than state of the art architectures. This project is more or less a verification of this claim by choosing target device as RPi.

To summarize, the goal of this project is to search for an energy efficient architecture while having maximum accuracy on RPi using FBNet with updated loss function equation \ref{newOptimization}. Baselines are crucial to compare our FBNet implementation performance. Even though architectures like ResNet-50 and VGG-16 have good accuracy they consume a lot of energy on RPi as they are not designed for ARM devices. Literature shows that CondenseNet \cite{huang2018condensenet} by Guo et al. and MobileNetV2 by Mark et al. stand out as state of the art architectures with best performance on ARM devices and hence they were chosen as baselines.

To calculate the actual baseline metrics we utilized the GitHub repositories provided by authors of MobileNetV2 and Condensenet as the source to preserve originality in implementation and ensure fault-tolerance.
The final details about benchmarking are as follows:

Datasets      : CIFAR 10\\
Architectures : CondenseNet-74 and MobileNetV2\\
Benchmark metrics    : Accuracy, Time per inference and Energy per inference

Accuracy of the models are mostly architecture agnostic, therefore they were tested on CPU based on TA's suggestion. Energy and latency per inference were profiled using techniques mentioned in previous section on the target device RPi, unlike accuracy. Final benchmarking results are shown in table below:

 \begin{table}[!h]
\centering
\caption{Benchmarking Results}
\resizebox{\columnwidth}{!}{\begin{tabular}{|c|c|c|c|c|c|c|c|c|c|c|}
\hline
\multirow{3}{*}{} & \multicolumn{5}{|c|}{CondenseNet} &  \multicolumn{5}{|c|}{MobileNetV2} \\ 
\hline
& \multicolumn{2}{|c|}{Accuracy(\%)} & \multicolumn{2}{|c|}{Latency(s)} & Energy(J) & \multicolumn{2}{|c|}{Accuracy(\%)} & \multicolumn{2}{|c|}{Latency(s)} & Energy(J) \\ 
\hline
& RPi & CPU & RPi & CPU & RPi & RPi & CPU & RPi & CPU & RPi \\ 
\hline
CIFAR-10 & 90.0 & 90.0 & 4.83 & 0.0665 & 9.83 & 92.0 & 92.0 & 7.1 & 0.0437 & 18.24\\ 
\hline
\end{tabular}
}
\label{benchmark}
\end{table}
Table \ref{benchmark} provides an idea as to how baseline models perform on our target device. We can see Mobilenetv2 achieves better accuracy over Condensenet with higher energy and latency. This project aims at finding an architecture that would be more energy efficient of the baselines while maximizing accuracy as possible, which is not possible when the architecture is fixed, like in the case of our baselines.

\subsection{NAS implementation}
\subsubsection{FBNet Basic Code Structure}
\begin{itemize}
    \item Candidate blocks - This code maintains a list of candidate blocks to be chosen from, for the 22 TBS layers. It is explained further below in Section \ref{cand}.
    \item SuperNet trainer - This code is the heart of our implementation and implements the main FBNet architecture and specifies its training process. It is explained further below in Section \ref{supernet}.
    \item Main program - It will get all the configuration parameters and hyperparameters and call the FBNet class. It will initiate 'train()' and 'search' functions and run them for the supernet. It will also keep a log of the proabilities in a file for post-training analysis.
    \item Childnet train program - Once the supernet is trained, a childnet is sampled from the optimized probability distribution and is trained for 120 epochs. This function is implemented in this code, which will take in the probability file from the main program.
    \item Utils file - This file implements helper functions like cosine decay for temperature.
    
\end{itemize}

\subsubsection{FBNet Candidate Block Implementation}\label{cand}

We implemented the fundamental unit of FBNet architecture, namely, FBNet Block built on MobileNetv2 architecture, in PyTorch. There are two main components to it:

\textbf{1.} Configurable FBNet block - This block takes in as inputs the following parameters: input channels, output channels, stride, kernel size, group size, expansion ratio. We implement the blocks as shown in Fig. \ref{blockFBnet} as a PyTorch sequential module containing pointwise Conv2D, ReLU, depthwise Conv2D, ReLU and pointwise Conv2D in the respective order. If group size = 2, channel shuffling is added after pointwise convolutions.

\textbf{2.} Candidate blocks - A list of blocks for FBNet 'TBS' blocks is created by calling the above configurable FBNet block with the appropriate list of parameters for the corresponding TBS block.
    
    The search process optimizes on latency and energy values read from their respective lookup tables. The latency and energy text files are generated for all blocks for all the 22 searchable layers. The ninth block in a layer would be the identity layer, a skip connection, that consumes the lowest energy and the least time. It can be present only if the number of input and output channels are equal for a layer.
    

\subsubsection{SuperNet Trainer Implementation}\label{supernet}
As mentioned above, we modularised our code into various sections of which supernet implementation is one of the most cardinal aspects. Here, we have implemented the fixed macro-architecture of FBNet with 24 layers, out of which the 22 TBS layers choose from the 9 candidate blocks mentioned above. We maintain the two trainable parameters 'weight' and 'theta' as \textit{nn.Parameter} in PyTorch. There are several important components to this code:
\begin{itemize}
    \item SuperNet structure - The supernet is basically a concatenated representation of the entire search space. The \textit{init} function creates the macro-architecture as shown in Fig. \ref{macroarch} by appending the corresponding fixed layers before and after the operator list for TBS layers (invoked from 'candidate blocks' code) and creating an \textit{nn.Sequential} structure out of it. It also appends a linear classifier at the end which corresponds to the fully connected layer. It reads in latency and energy files and creates PyTorch tensors for both.\\The \textit{forward} function defines the forward flow of data through the structure created above. It applies a gumbel softmax function for probabilities using 'theta's and uses it to (probabilistically) add up the latency and energy values for all blocks for all layers. This gives latency and energy loss functions differentiable with respect to 'theta'. Adding these losses to the cross-entropy loss defines the final loss function.
    \item Search process and weight/theta update - The search process starts with weights initialized randomly and probabilities initialized to 1. Our implementation takes in epochs and batch size as command-line arguments among others. We follow FBNet paper for implementing the search process. Our supernet is trained for 90 epochs with a batch size of 256. We provide a warmup of 10 epochs to stabilize the weights before beginning to train 'theta'. After the first 10 epochs, in every epoch, weights are updated on 80\% training data using SGD with momentum and theta is then updated on the rest 20\% using Adam optimizer. We have used the same initial learning rate and decay values (weight decay, temperature decay) as in FBNet.
    \item Result logging - This code displays the current values of different losses (cross-entropy, latency, energy) and accuracy after every epoch of training 'weight' and 'theta'. It also saves the 'theta' values in a file after every epoch. The latest 'theta' file can be used to sample the child architecture.
\end{itemize}



\section{Results}
\subsection{Evaluation}
The child neural architectures discovered during a FBNAS search are termed as FBNets. However, since in our case we have an updated loss function (with energy term), we call the childnets as our HANNA models. We trained 30 different HANNA models for RPi target device to check for their performance over baseline architectures. The evaluation procedure is as follows:

\textbf{1.} Profile the RPi to create latency and energy lookup tables.\\
\textbf{2.} Choose required values of alpha, beta, gamma, delta for model \textit{n} which are collectively referred to as \textit{knobs}.\\
\textbf{3.} Visualize the effect of these values on initial latency and energy values. We call the visualized values as vLAT and vENER. Calculate vENER/vLAT and vLAT/vENERGY which are the energy dominance and latency dominance metrics respectively.\\
\textbf{4.} Train the updated FBNET model with chosen \textit{knob} values for a total of 90 epochs which includes a warmup of 10 epochs. We also pass arguments referring to path of the directory to store logs.\\
\textbf{5.} Record the values of latency, energy, accuracy and total loss of the childnet converged at the last epoch.\\
\textbf{6.} Parse the probabilities file to infer the childnet (HANNA) which is then trained on GPU for 120 epochs.\\
\textbf{7.} Record the accuracy of childnet after training.\\
\textbf{8.} Use the inferred architecture to measure latency per inference and energy per inference on target RPi.\\
\textbf{9.} Repeat the procedure for 29 other models. Profiling step is not repeated.

\begin{figure}[!h]
    \centering
    \frame{\includegraphics[width=0.48\textwidth]{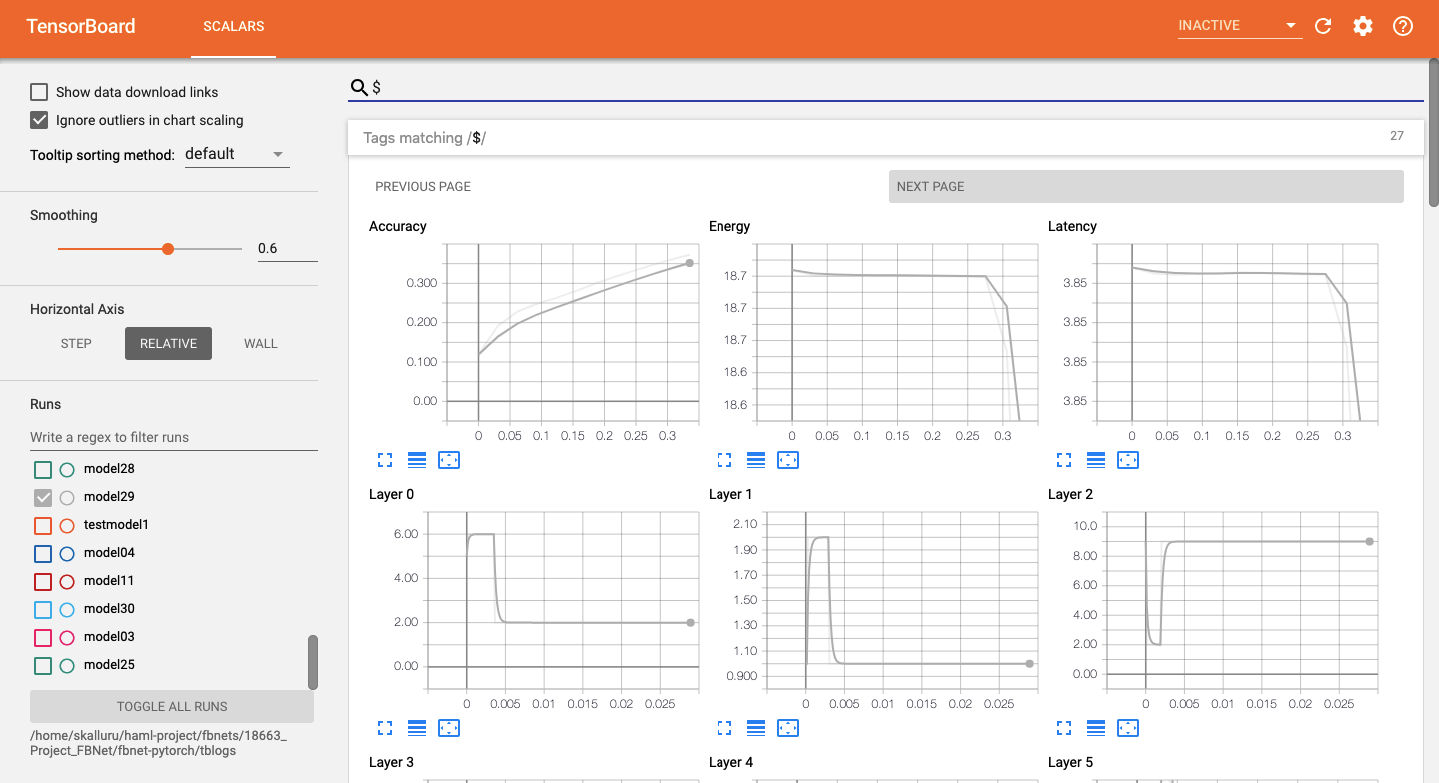}}
    \caption{Tensorboard visualization for best HANNA model. Note the increase in accuracy and drop in latency and energy over increasing epochs.}
    \label{tensorboard}
\end{figure}

The goal of evaluation is to observe the effect of different \textit{knobs} on the search process. As a result, we have chosen alpha, beta, gamma and delta such that half the models are tuned to be ENERGY DOMINANT and the other half to be LATENCY DOMINANT. Before tuning the model, it is important to pre-visualize how the knobs perform, therefore we define simple variables like vENER and vLAT which would pick initial values of latency and energy, apply alpha and beta (for vLAT), gamma and delta (for vENER) and determine their effect. In our case, the initial latency value of the architecture was equal to 3.5s and energy equal to 18.43J, for knob values of alpha=0.5, beta=1.5, gamma=0.5, delta=0.5 we would have $vLAT = 0.5 * (3.5)^{1.5}$ and $vENER = 0.5*(18.43)^{0.5}$. vLAT and vENER help us visualize how the knob values would affect the loss equation, a larger vLAT implies the loss equation penalises more for LAT(a) than for ENER(a), which encourages the model to strive harder and find architectures focussed more on minimizing latency loss. The same concept applies for vENER as well.  
vENER and vLAT determines whether the model was tuned to be latency or energy dominant, if vENER/vLAT ratio is greater than 1.0, then it is tuned to be energy dominant else latency dominant. We use tensorboard to visualize the accuracy, latency and energy of each new childnet sampled at every epoch as shown in Fig. \ref{tensorboard}.

\subsection{Inference}


\begin{figure}[!t]
    \centering
    \frame{\includegraphics[width=0.5\textwidth]{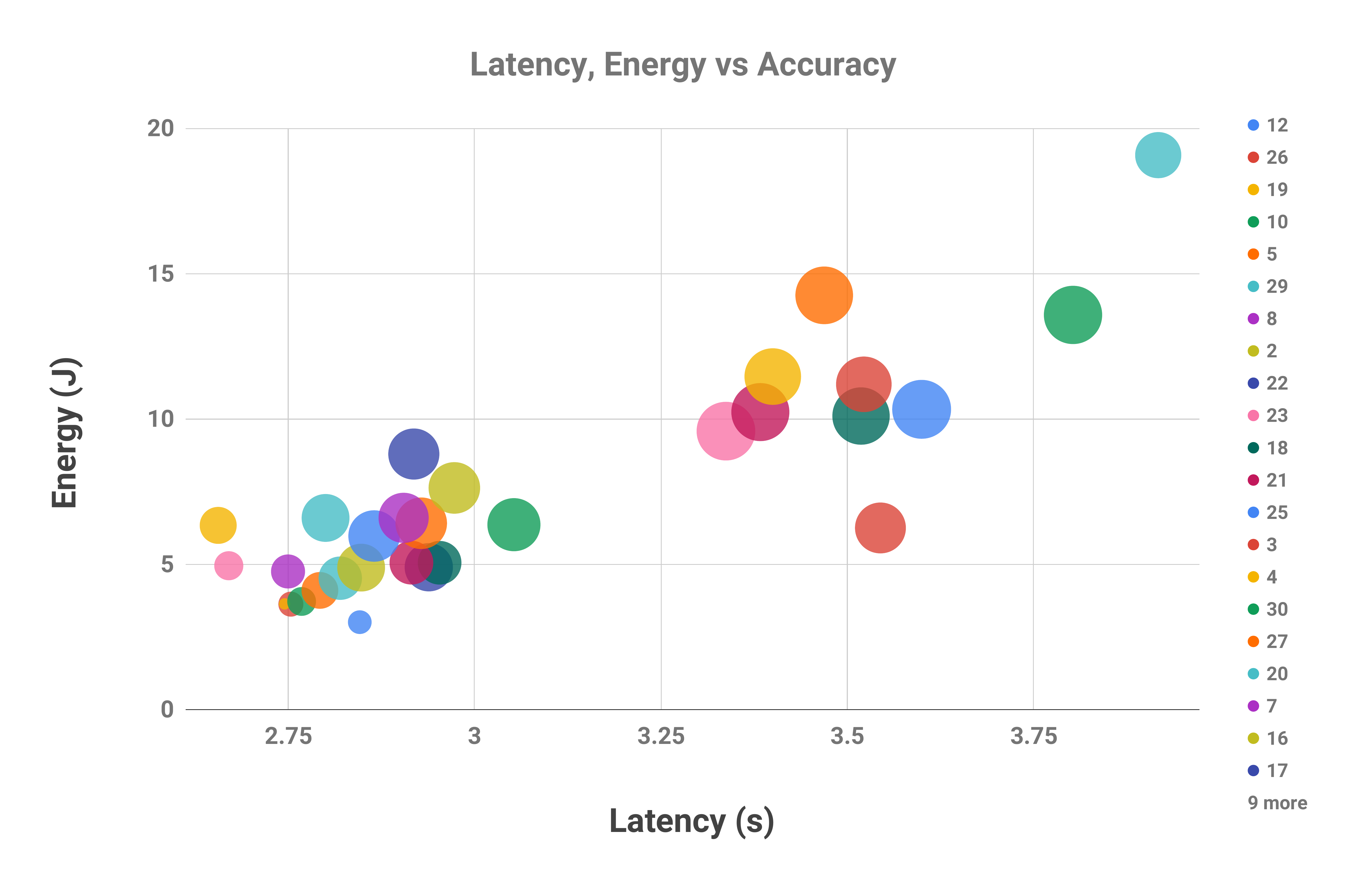}}
    \caption{Latency vs Energy vs Accuracy - Every bubble represents a different model and size of bubble is proportional to its corresponding accuracy. Closer to origin with larger size is better.}
    \label{lat_ener_acc}
\end{figure}
Fig. \ref{lat_ener_acc} shows the multi-objective trade-off between accuracy, energy and latency for all our 30 child models (HANNAs). We can see that as a general high-level trend, the models that are further apart from the origin (higher energy and latency) are larger in size (higher accuracy) and vice versa. Model 15 (purple bubble towards the right) displays highest accuracy (91.2\%) whereas model 4 (yellow bubble to the extreme left) and model 12 (blue bubble at the bottom) provide lowest latency (2.65 s) and energy respectively (3 J). Then there are models in between that provide various levels of trade-offs. All these models which either maximize (or minimize) accuracy (or latency or energy) for a fixed value of other metrics are Pareto-optimal. Any of the child architectures that fall on such Pareto curves can be argued as optimal for our target device.

We chose a model (model 29 - fairly sized blue bubble close to origin) whose latency and energy values fall within 15\% of the minimum values, still maintaining a decent accuracy of 87.7\%. Its metrics are shown in Table \ref{tab:final_results} against that of state-of-the-art mobile/embedded neural architectures, namely, MobileNetv2 and CondenseNet. We can see that HANNA (model 29) found by our NAS outperforms them on Raspberry Pi by 2.5x and 1.7x in terms of latency, and 3.8x and 2x in terms of energy consumption respectively, at the cost of a fairly tolerable accuracy drop of 4-5\%.

\begin{figure}[!t]
    \centering
    \frame{\includegraphics[width=0.5\textwidth]{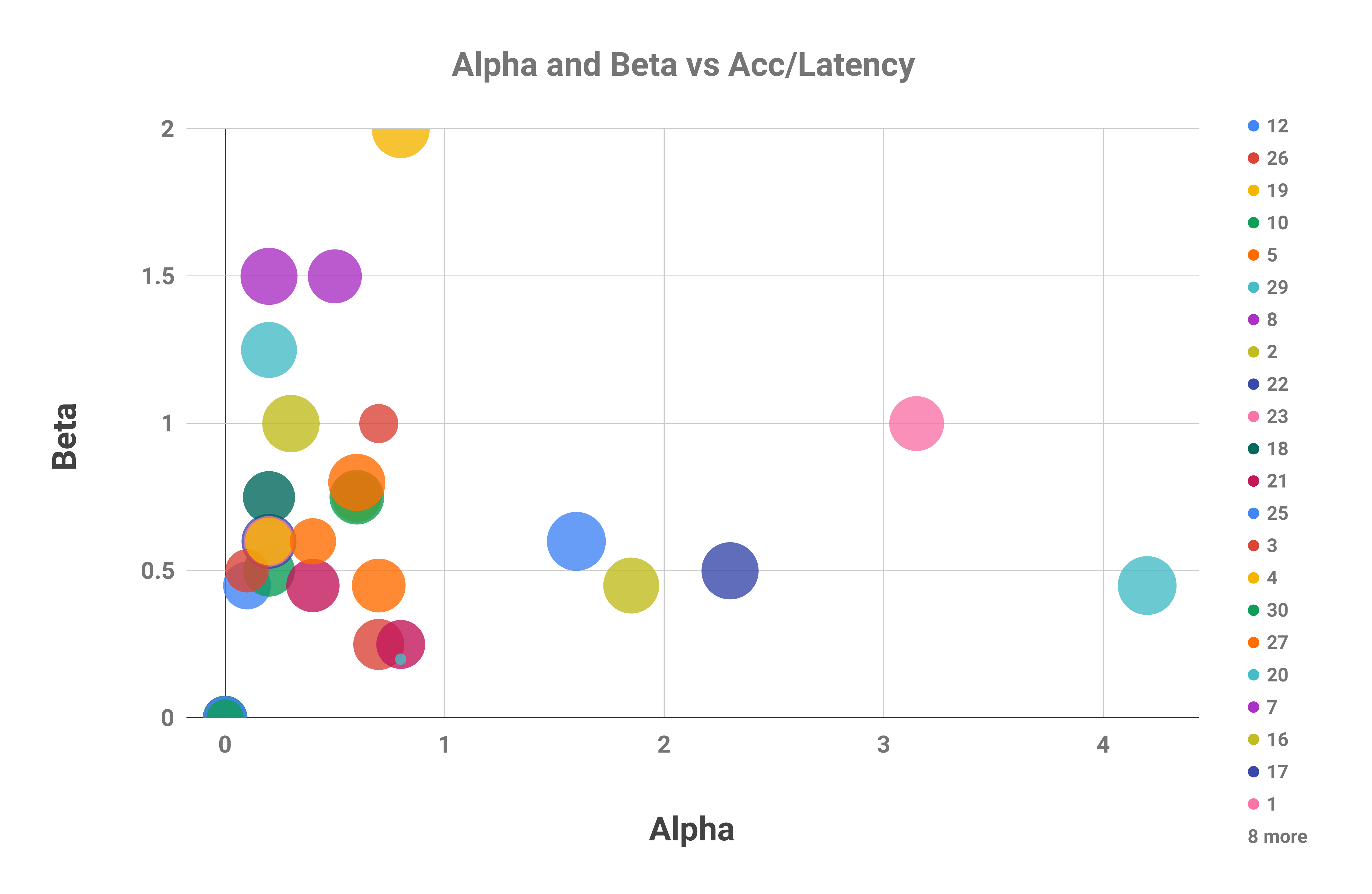}}
    \caption{Alpha and Beta vs Accuracy/Latency - Every bubble represents a different model and size of bubble is proportional to its corresponding (accuracy/latency) value. Larger size is better.}
    \label{ab_acc_lat}
\end{figure}

Fig.\ref{ab_acc_lat} represents how the variation in $\alpha$ and $\beta$ affects the accuracy/latency ratio. The circles in Fig.\ref{ab_acc_lat} represents the accuracy/latency ratio, the larger the ratio the greater is the size of the bubble. The goal of this experiment was to produce highly accurate and low latent models. An increase in $\alpha$ brings about an improvement in accuracy/latency as the bubbles shift towards to the right. Model 29 in Fig.\ref{ab_acc_lat} has the best accuracy/latency ratio of 23.9. The trend shows that as $\alpha$ increases while $\beta$ is constant the accuracy  improves slightly while the latency stays around the same.

\begin{figure}[!t]
    \centering
    \frame{\includegraphics[width=0.5\textwidth]{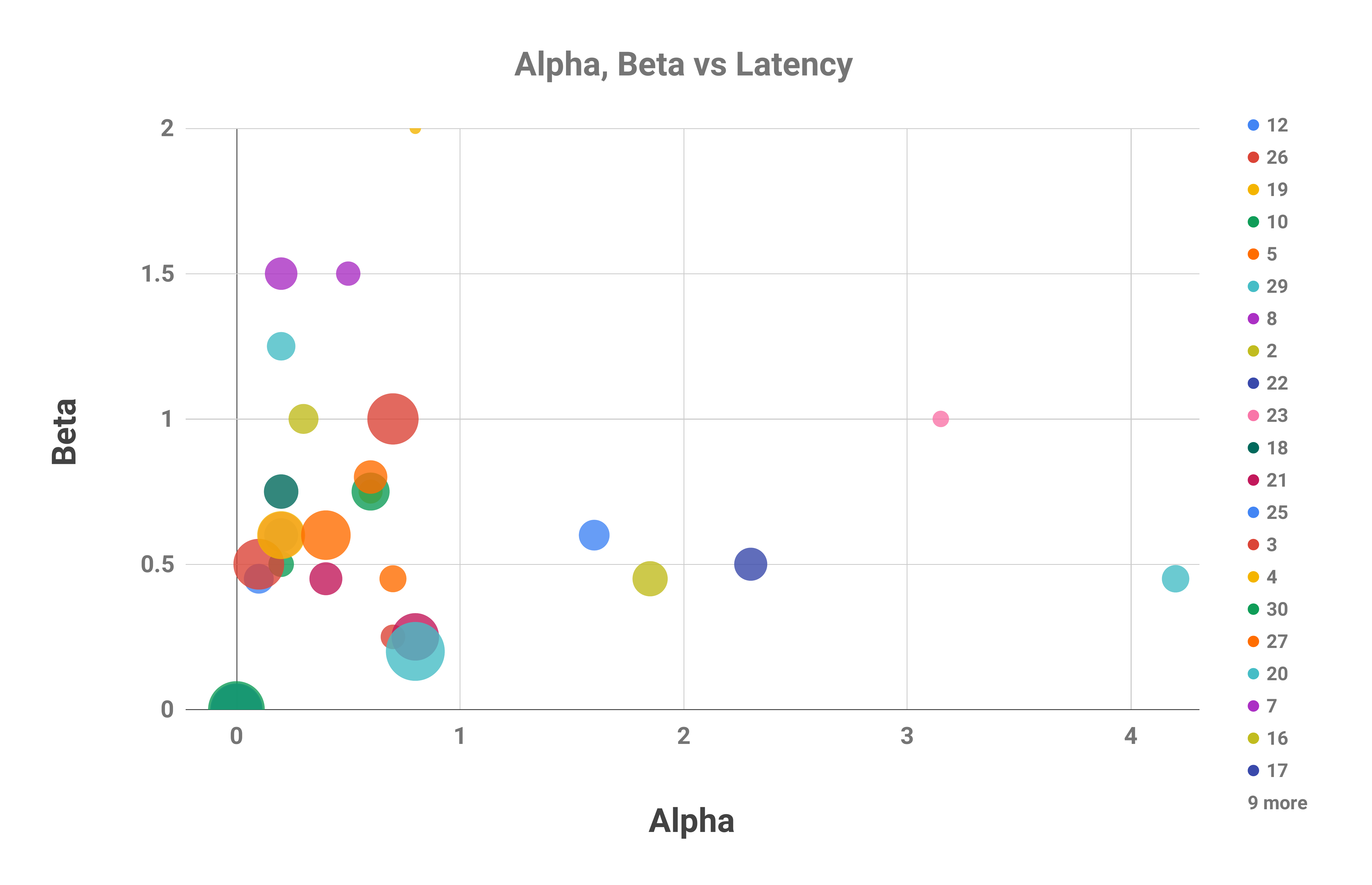}}
    \caption{Alpha and Beta vs Latency - Every bubble represents a different model and size of bubble is proportional to its corresponding latency. Smaller size is better.}
    \label{ab_lat}
\end{figure}

Fig. \ref{ab_lat} plots $\alpha$ and $\beta$ (which determine the contribution from latency part of the loss function) against latency for all 30 child HANNAs. Latency is represented by the size of bubbles in this plot. We can see that, as expected, when $\alpha$ and $\beta$ is increased thereby increasing the dominance of latency relative to energy in the loss function, the final latency of the model actually decreases. In other words, if we move farther from the origin in fig. \ref{ab_lat}, the size of the bubble decreases as a general trend. An interesting observation here is that models 4 (orange bubble at the very top) and 23 (pink bubble towards top right) which are some of the farthest bubbles from the origin have the lowest latencies (about 2.67 s). However, model 20 (towards bottom right) which is almost as far from the origin as model 4 (towards top left) has a higher latency than model 4. This confirms that changing $\beta$ (exponential factor) is more effective than changing $\alpha$ (multiplicative factor) in reducing latency.

\begin{figure}[!t]
    \centering
    \frame{\includegraphics[width=0.5\textwidth]{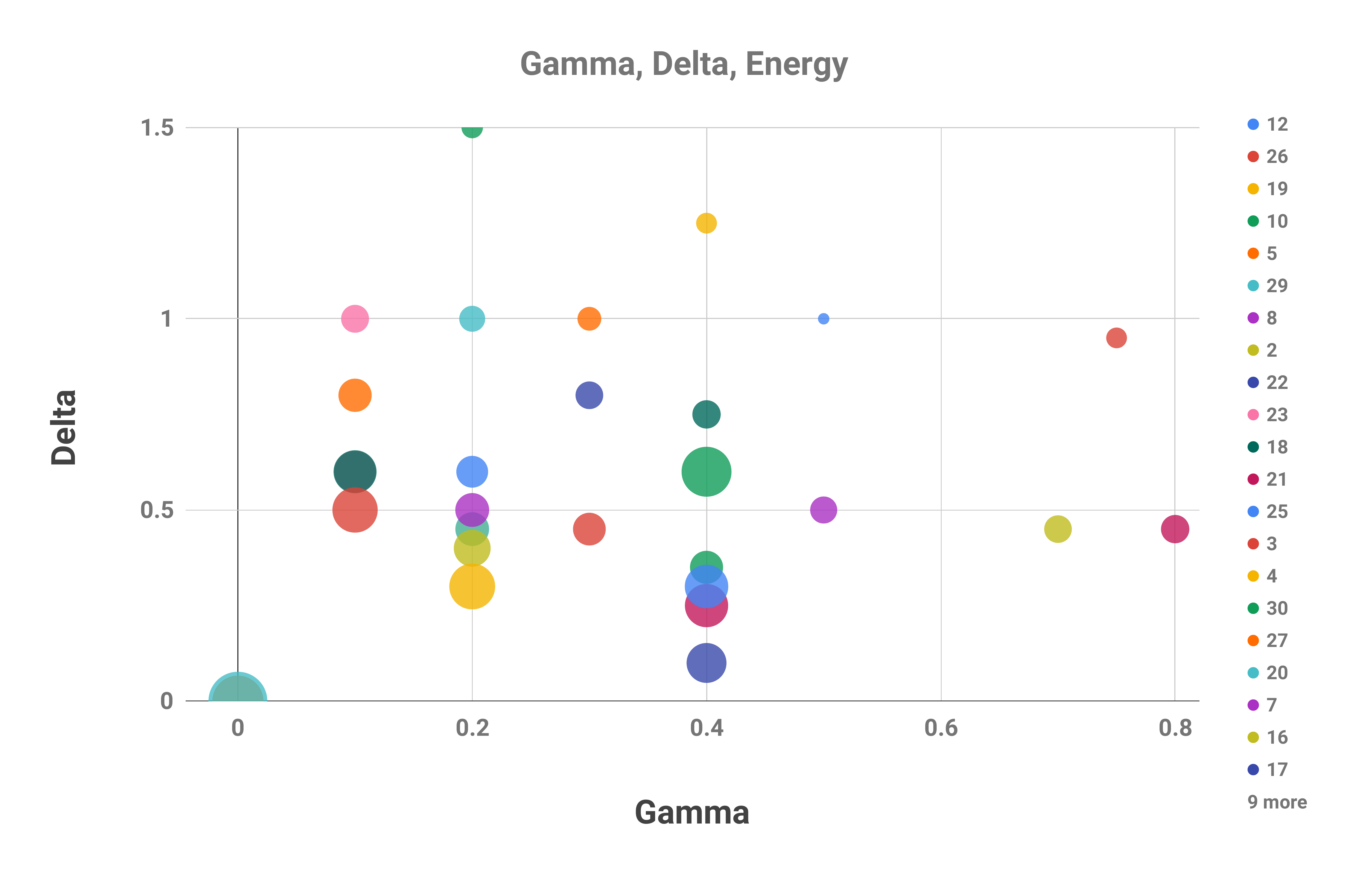}}
    \caption{Gamma and Delta vs Energy - Every bubble represents a different model and size of bubble is proportional to its corresponding energy. Smaller size is better.}
    \label{gd_ener}
\end{figure}

Fig. \ref{gd_ener} shows a similar plot but for $\gamma$ and $\delta$ versus energy consumption. A complementary interesting observation that can be made from this plot is that for a particular value of $\gamma$, as we increase $\delta$, energy of the models clearly decrease (models shrink in size as we go up for a particular $\gamma$). Model 12 (small blue bubble which is also one of the farthest from origin) has the lowest energy consumption (about 3 J).\\
We define latency dominance metric as the ratio of initial values of latency and energy losses weighted using $\alpha$, $\beta$, $\gamma$ and $\delta$. This metric indicates the skew of the loss function towards latency optimization over energy optimization, given that particular configuration of $\alpha$, $\beta$, $\gamma$ and $\delta$. Similarly, we define energy dominance metric as the inverse of latency dominance metric.

\begin{table}[!h]
\caption{Summary of final results}
\label{tab:final_results}
\begin{tabular}{|c|c|c|c|}
\hline
     & Accuracy (\%) & Latency (s) & Energy (J) \\ \hline
  MobileNetv2 & 92.4 & 7.1 & 18.24 \\ \hline
  CondenseNet & 91.1 & 4.83 & 9.28 \\ \hline
  HANNA & 87.7 & 2.88 & 4.79 \\ \hline
\end{tabular}

\end{table}

\subsection{Comparison with baselines in realtime}

To aid in the inference process, a utility script was implemented to sample a childnet model based on the most probable theta value per layer. The generated childnet was trained on a GPU for 120 epochs and then inferred for accuracy on a laptop CPU. It was then deployed on the RPi to check for energy and latency performance.

Based on the results in Table.\ref{tab:final_results}, HANNA performs significantly better than CondenseNet and MobileNetV2 in terms of both energy and latency. The energy and latency metrics for HANNA is around 4x and 2.46x better than MobileNetV2 while around 1.93x and 1.67x better than CondenseNet.

HANNA is optimized to produce energy and latency aware models while still maximizing accuracy. The accuracy of HANNA is within a reasonable threshold of the two state of the art models and is therefore acceptable given that it is optimized for three factors:accuracy, latency, and energy.

\section{Conclusion}

Knobs like alpha, beta, gamma and delta help us to create well optimized models for required target device and application. Let us consider an object detection application for for self driving application. Here, low-inference time models are preferred for which we can tune models to be latency dominant. The same task for a different target device like drone delivery would need our model to less-energy consuming but latency can be compromised. For such an application we can tune gamma and delta to create energy-dominant models. 

\begin{figure}
    \centering
    \includegraphics[width=0.4\textwidth]{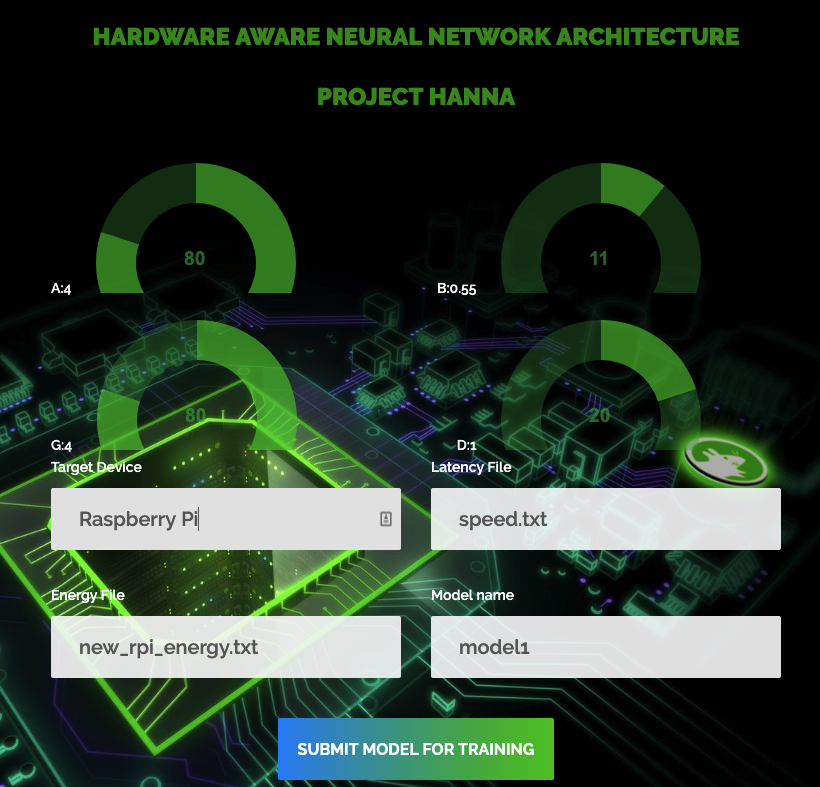}
    \caption{Webpage for creating HANNA models}
    \label{webpage}
\end{figure}

In future, we believe that tuning model for hardware architectures would be available like a web API, where an user can specify how much does he want to tune the model towards latency and energy. The user would also provide the lookup tables for the target device. The API would train the model online and provide the theta file for the user to download. The user can later train HANNA and directly deploy the model on the target device, all of this requires no hardware or deep learning knowledge from the users side. Inspired by this idea, we created a webpage (Fig. \ref{webpage}) which performs takes in alpha,beta,gamma and delta values along with appropriate lookup tables, performs NAS search on its backend GPU and provides with theta file at the end of training. Our implementation can be found on Github at: \textit{\href{https://github.com/hpnair/18663\_Project\_FBNet}{\color{blue}https://github.com/hpnair/18663\_Project\_FBNet}}

\bibliography{references_hanna}

\begin{thebibliography}{6}
\providecommand{\natexlab}[1]{#1}
\providecommand{\url}[1]{\texttt{#1}}
\expandafter\ifx\csname urlstyle\endcsname\relax
  \providecommand{\doi}[1]{doi: #1}\else
  \providecommand{\doi}{doi: \begingroup \urlstyle{rm}\Url}\fi

\bibitem[Hsu et~al.(2018)Hsu, Chang, Juan, Pan, Chen, Wei, and
  Chang]{hsu2018monas}
Hsu, C.-H., Chang, S.-H., Juan, D.-C., Pan, J.-Y., Chen, Y.-T., Wei, W., and
  Chang, S.-C.
\newblock Monas: Multi-objective neural architecture search using reinforcement
  learning.
\newblock \emph{arXiv preprint arXiv:1806.10332}, 2018.

\bibitem[Huang et~al.(2018)Huang, Liu, Van~der Maaten, and
  Weinberger]{huang2018condensenet}
Huang, G., Liu, S., Van~der Maaten, L., and Weinberger, K.~Q.
\newblock Condensenet: An efficient densenet using learned group convolutions.
\newblock In \emph{Proceedings of the IEEE Conference on Computer Vision and
  Pattern Recognition}, pp.\  2752--2761, 2018.

\bibitem[Liu et~al.(2018)Liu, Simonyan, and Yang]{liu2018darts}
Liu, H., Simonyan, K., and Yang, Y.
\newblock Darts: Differentiable architecture search.
\newblock \emph{arXiv preprint arXiv:1806.09055}, 2018.

\bibitem[Mingxing~Tan \& Le(2018)Mingxing~Tan and Le]{tan2018mnas}
Mingxing~Tan, Bo~Chen, R. P. V.~V. and Le, Q.~V.
\newblock Mnasnet: Platform-aware neural architecture search for mobile.
\newblock \emph{arXiv preprint arXiv:1807.11626}, 2018.

\bibitem[Sandler et~al.(2018)Sandler, Howard, Zhu, Zhmoginov, and
  Chen]{sandler2018mobilenetv2}
Sandler, M., Howard, A., Zhu, M., Zhmoginov, A., and Chen, L.-C.
\newblock Mobilenetv2: Inverted residuals and linear bottlenecks.
\newblock In \emph{Proceedings of the IEEE Conference on Computer Vision and
  Pattern Recognition}, pp.\  4510--4520, 2018.

\bibitem[Wu et~al.(2018)Wu, Dai, Zhang, Wang, Sun, Wu, Tian, Vajda, Jia, and
  Keutzer]{wu2018fbnet}
Wu, B., Dai, X., Zhang, P., Wang, Y., Sun, F., Wu, Y., Tian, Y., Vajda, P.,
  Jia, Y., and Keutzer, K.
\newblock Fbnet: Hardware-aware efficient convnet design via differentiable
  neural architecture search.
\newblock \emph{arXiv preprint arXiv:1812.03443}, 2018.

\end{thebibliography}
\bibliographystyle{sysml2019}

\section{Acknowledgement}
We would like to sincerely thank Prof. Diana Marculescu and Dr.Da-Cheng Juan for their guidance and support in the project. We would also like to thank Angie Cheng for discussing about DPPNets and hypervolume based loss function with us.

\end{document}